# Defining and Generating Axial Lines from Street Center Lines for better Understanding of Urban Morphologies

Xintao Liu and Bin Jiang

Department of Technology and Built Environment, Division of Geomatics
University of Gävle, SE-801 76 Gävle, Sweden
Email: xintao.liu@hig.se, bin.jiang@hig.se

*(Draft: September 2010, Revision: April 2011)*

**Abstract**
Axial lines are defined as the longest visibility lines for representing individual linear spaces in urban environments. The least number of axial lines that cover the free space of an urban environment or the space between buildings constitute what is often called an axial map. This is a fundamental tool in space syntax, a theory developed by Bill Hillier and his colleagues for characterizing the underlying urban morphologies. For a long time, generating axial lines with help of some graphic software has been a tedious manual process that is criticized for being time consuming, subjective, or even arbitrary. In this paper, we redefine axial lines as the least number of individual straight line segments mutually intersected along natural streets that are generated from street center lines using the Gestalt principle of good continuity. Based on this new definition, we develop an automatic solution to generating the newly defined axial lines from street center lines. We apply this solution to six typical street networks (three from North America and three from Europe), and generate a new set of axial lines for analyzing the urban morphologies. Through a comparison study between the new axial lines and the conventional or old axial lines, and between the new axial lines and natural streets, we demonstrate with empirical evidence that the newly defined axial lines are a better alternative in capturing the underlying urban structure.

**Keywords:** Space syntax, street networks, topological analysis, traffic, head/tail division rule

## 1. Introduction

Axial lines refer to the longest visibility lines for representing individual linear spaces in urban environments. The least number of axial lines cutting across the free space of an urban environment constitute what is often called an axial map. Based on the axial map, or more precisely the intersection of the axial lines, space syntax (Hillier and Hanson 1984, Hillier 1996) adopts a connectivity graph consisting of nodes representing axial lines and links if the axial lines are intersected for understanding the underlying morphology. The status of individual nodes in the connectivity graph can be characterized by some defined space syntax metrics, most of which have a closed link to graph theoretic metrics such as centrality (Freeman 1979). Thus all axial lines are assigned some metrics for characterizing their status within the axial map. This is fundamental to space syntax for urban morphological analysis. However, for a long time generating the axial lines has been a tedious manual process using some GIS or CAD software tools. It is considered to be time consuming, subjective, or even arbitrary (e.g., Jiang and Claramunt 2002, Ratti 2004). Researchers have attempted to seek automatic solutions (e.g., Peponis et al. 1998, Turner, Penn and Hillier 2005, Jiang and Liu 2010a), but none of them really work efficiently and effectively for large cities.

We believe that the lack of an automatic solution to generating axial lines for large cities is due to the ambiguity of the conventional definition of axial lines. The conventional definition is essentially based on the notion of visibility. This definition works well for small urban environments in which buildings are visual obstacles and the space between the buildings constitutes the free space on which people can freely move around. For example, the automatic solutions mentioned above are mainly targeted for such a case. However, we cannot adopt the same definition at the city level for generating the axial lines. This is because human movement at the city level is constrained by street networks rather than the free space between buildings. We therefore think that visibility cannot simply be applied to generating axial lines for large cities. Instead, we suggest walkability or drivability as the basic notion for defining the axial lines at the city level. We define axial lines as the least number of individual straight line segments mutually intersected along natural streets that are generated from street center lines using the Gestalt principle of good continuity (Jiang, Zhao and Yin 2008, Thomson 2003). Based on the new definition, we can develop an automatic solution to generating the axial lines from street center lines.

This paper is further motivated by the increasing availability of volunteered geographic information contributed by individuals and supported by the Web 2.0 technologies (Goodchild 2007). In this respect, the OpenStreetMap (OSM, www.openstreetmap.org) community has collected and contributed over a few hundred gigabytes of street network data for the entire world. The quality and quantity of the OSM data for Europe and North America can be compared to that of the data maintained by the national mapping agencies. We believe that the volunteered geographic information or the OSM data in particular provides an unprecedented data source for various urban studies. Furthermore, they can successfully be used to generate axial maps for individual cities.

The contribution of this paper is three-fold: (1) we provide a new definition of axial lines, and consequently an automatic solution to generating the axial lines from street center lines, (2) we conduct a comparison study between the new axial lines and the conventional axial lines and between the new axial lines and the natural streets, and find that the new axial lines can be a better alternative for illustrating the underling urban morphologies, and (3) along with the first point, we demonstrate to our surprise an application of the head/tail division rule. This rule illustrates a regularity that can be used for partitioning values (that exhibit a heavy tailed distribution) between a minority head and a majority tail; refer to (Jiang and Liu 2010c) for more details.

The remainder of this paper is structured as follows. In Section 2, we re-define the axial lines and introduce a procedure to automatically generate them from street center lines. This new definition is justified from the limitation of the conventional definition initially developed by Bill Hillier and his co-workers. In Section 3, we apply the procedure to six street networks, and generate the new axial maps and compute the related space syntax metrics for illustrating the underlying urban morphologies. We discuss some results from a comparison study between the new axial lines and old axial lines, and between the new axial lines and natural streets in terms of how they capture the underlying urban morphologies and traffic flow. Finally, Section 5 provides a conclusion to this paper and points out future work.

## 2. Re-defining and auto-generating the axial lines
Before re-defining the axial lines, let us take a look at how they were conventionally defined. The initial definition of the axial lines is based on a prior definition of the convex map of free space (Hillier and Hanson 1984). The convex map is defined as the least set of fattest spaces that covers the free space. Based on the prior definition, axial lines are defined as '*the least set of such straight lines which passes through each convex space and makes all axial links*' (p. 92). In practice, no one seems to care about the definition of a convex map. Instead a simple procedure is adopted for generating the axial lines, i.e., '*first finding the longest straight line that can be drawn . . ., then the second longest, and so on until all convex spaces are crossed and all axial lines that can be linked to other axial lines without repetition are so linked*' (p. 99). This conventional definition of axial lines or the procedure of generating axial lines relies much on the notion of visibility. Due to this fact, the axial lines are also called the longest visibility lines. This definition makes perfect sense for a small urban environment in which buildings are considered visual obstacles and the space between buildings is walkable. In this circumstance, there are already some automatic solutions to generating the axial lines (Jiang and Liu 2010a, Turner, Penn and Hillier 2005, Peponis et al. 1998).

For large cities, the space between buildings is not always walkable and only streets or sidewalks are truly walkable. Thus we cannot rely on the visibility between buildings to generate axial lines. Instead, we must consider only the walkable space for generating axial lines. In this paper we define axial lines as the least number of straight lines that are mutually intersected along individual natural streets. A natural street is defined as a self-organized street generated from individual adjacent street segments which have the smallest deflection angles. Perceptually, the self-organized natural street form a good continuity based on the principle of Gestalt psychology (Jiang, Zhao and Yin 2008, Thomson 2003). Under this definition, two parallel straight streets with a few meters gap in between, yet visible to each other, would be represented as two axial lines, since they are distinctly walkable or drivable spaces. The same idea is applied to a highway that is separated by a small barrier into two different driving lanes. Eventually, the least number of the longest axial lines constitute the newly defined axial map.

Based on this new definition, generating axial lines becomes a relatively easy and straightforward task. First, we need to form individual natural streets based on street segments or arcs and using some join principles such as every-best-fit, self-best-fit and self-fit (Jiang, Zhao and Yin 2008). Among the three join principles, the every-best-fit principle is the best choice, since it tends to form natural streets that are similar or close to the corresponding named streets (Jiang and Claramunt 2004). As suggested in the previous study, we adopt 45 degrees as the threshold angle for terminating the join process. Once the natural streets are generated, we then assess their curviness and convert them into a set of axial lines. If the initial natural streets are straight enough, they directly become axial lines. For those streets with a big bend, we chop them into two or several straight parts based on the degree or extent of curviness.

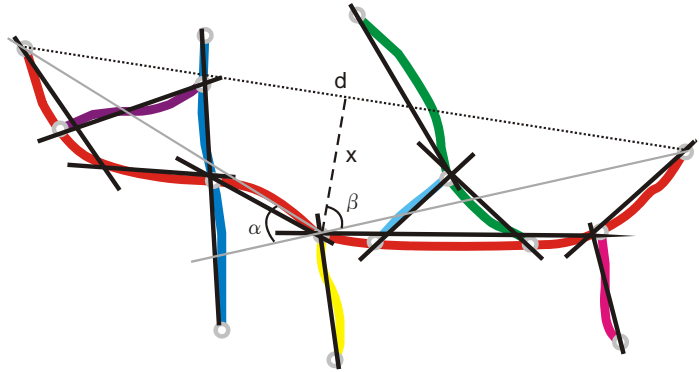

Figure 1: (Color online) Illustration of converting natural streets (color lines) to the final axial lines (black lines) (Note: dotted lines are the base line with distance d, while dashed lines indicate how far (x) the farthest vertices are from the base lines. The two gray lines are possible axial lines, but not the final ones)

To identify a big bend, we simply draw a base line linking two ending points of a natural street (with distance d), and check how far (x) the farthest vertex is to the base line (Figure 1). In fact, both x and x/d follow a lognormal distribution – one of the heavy tailed distributions; we will illustrate this fact in the following experiments. This fact allows us to use the head/tail division rule (Jiang and Liu 2010c) for the following chopping process. The head/tail division rule states that *given a variable V, if its values, v, follow a heavy tailed distribution, then the mean (m) of the values can divide all the values into two parts: a high percentage in the tail, and a low percentage in the head*. It is a bit surprising to us that the mean of *x* or *x/d* can make such perfect sense for the chopping process. The underlying idea of identifying bends for the chopping process is to some extent similar to the detection of character points in line simplifications or generalization (e.g., Douglas and Peucker 1973, Ramer 1972), but we introduce another parameter ratio *x/d*. More importantly, the thresholds for the parameters are automatically determined by the head/tail division rule in the process. This process can be summarized as the following recursive function:

Initialize mean(x) and mean(x/d)
Calculate x and x/d of current street
**Function Chop (x, x/d, current street)**
    **If** ((x > mean(x) **and** x/d >= 10%*mean(x/d)) **or** (x <= mean(x) **and** x/d >= mean(x/d)))
        Split current street into two pieces: back street and front street
        Calculate x and x/d of back street
        **Chop (x, x/d, back street)**
        Calculate x and x/d of front street
        **Chop (x, x/d, front street)**
    **Else**
        Link two ending points of the current street as an axial line

To better understand the above function, let us adopt some specific figures to elaborate on the chopping process. We learn from Table 1 that the mean of x is about 30 meters and that of x/d is about 15%. If x is greater than 30 AND x/d is greater than 1.5% (10% of 15%), then we will split a natural street into two. Alternatively, if x is less than 30 AND x/d is greater than 15%, then we will do the same chopping. Of course, the two parameters x and x/d vary from one network to another as we see in Table 1, but every network has its own parameter settings derived from its own mean for the chopping process.

Let us explain how the 1.5% or 10%*mean(x/d) in general is determined, and what the implication of this parameter setting is. For the sake of simplicity, let us assume that mean(x) = 30, and mean(x/d) = 15% again; refer to Table 1 for actual parameter settings. With reference to Figure 1, suppose the farthest vertex is in the middle of a street, we note that the two angles have the relationship, $2\beta + \alpha = 180$. We learn from basic trigonometry that $\tan(\beta) = \frac{d}{2x}$, so $\beta = \text{atan}\left(\frac{d}{2x}\right)$. Given x/d >=1.5%, $\beta = \text{atan}\left(\frac{100}{3}\right) = 88.28$, and $\alpha = 180 - 2 * 88.28 = 3.44$. Therefore the 1.5% parameter setting implies that all the deflection angles between intersected axial lines derived from one street are at least 3.44 degrees. In other words, two valid intersected axial lines must have deflection angles greater than 3.44. This criterion is used to cross-check those axial lines derived from different streets that still intersect each other. If the deflection angle between two intersected lines is less than 3.44, they are replaced by one axial line. This process is somewhat like generating continuity lines (Figueiredo and Amorim 2005). What is unique for our approach is that all axial lines are generated under the same condition, which is statistically determined.

Apart from the above procedure, we develop an additional function to detect roundabouts. Initially roundabouts are for road safety purposes, but structurally they serve the same purpose as street junctions. In this respect, we have to differentiate roundabouts from ring roads based on their sizes. It is also important to note that roundabouts are very common in European cities, but they hardly appear in North American cities.

**3. Experiments on generating the axial lines**
We choose six city street networks for the following experiments and these generated networks are adopted from a previous study (Jiang and Liu 2010b). The six cities reflect typical street patterns in the literature, representing different morphological structures (Jacobs 1995). The three North American cities are grid-like and planned, while the three European cities look irregular and self-evolved. All the street networks were downloaded from the OSM databases. The street networks are shown in Figure 2. We first create topological relationships for the street networks. This has to be done since the original OSM data are without topology, much like digitized lines without generated coverage – a topology-based vector data format. Through the process of creating topology, all line segments will be assigned a direction and become arcs that meet at nodes and have left and right polygons. This can be easily done with some GIS software packages such as ArcGIS. Next, based on the arcs based street networks, we generate natural streets according to the every-best-fit principle and parameter settings mentioned in Section 2; the algorithmic functions can be found in Jiang, Zhao, and Yin (2008). The natural streets are visualized using a spectral color legend with red and blue respectively representing the highest and lowest local integration (Figure 3). Local integration is one of space syntax metrics for characterizing integration or segregation of streets. The classification is based on Jenks' natural break (Jenks 1967), so the variation within classes is minimized, while the variation between classes is maximized.

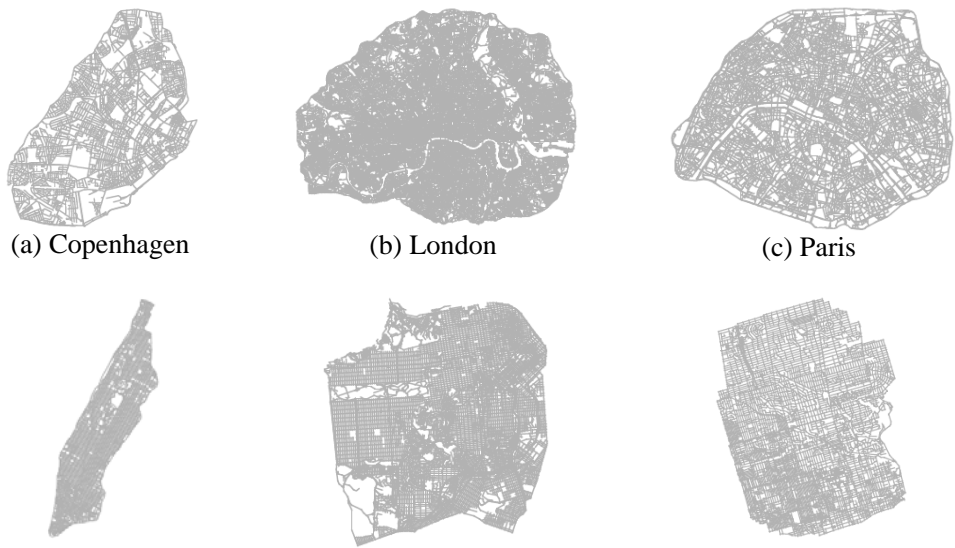

(a) Copenhagen      (b) London      (c) Paris

(d) Manhattan (e) San Francisco (f) Toronto

Figure 2: The six street networks (a) Copenhagen, (b) London, (c) Paris, (d) Manhattan, (e) San Francisco, and (f) Toronto

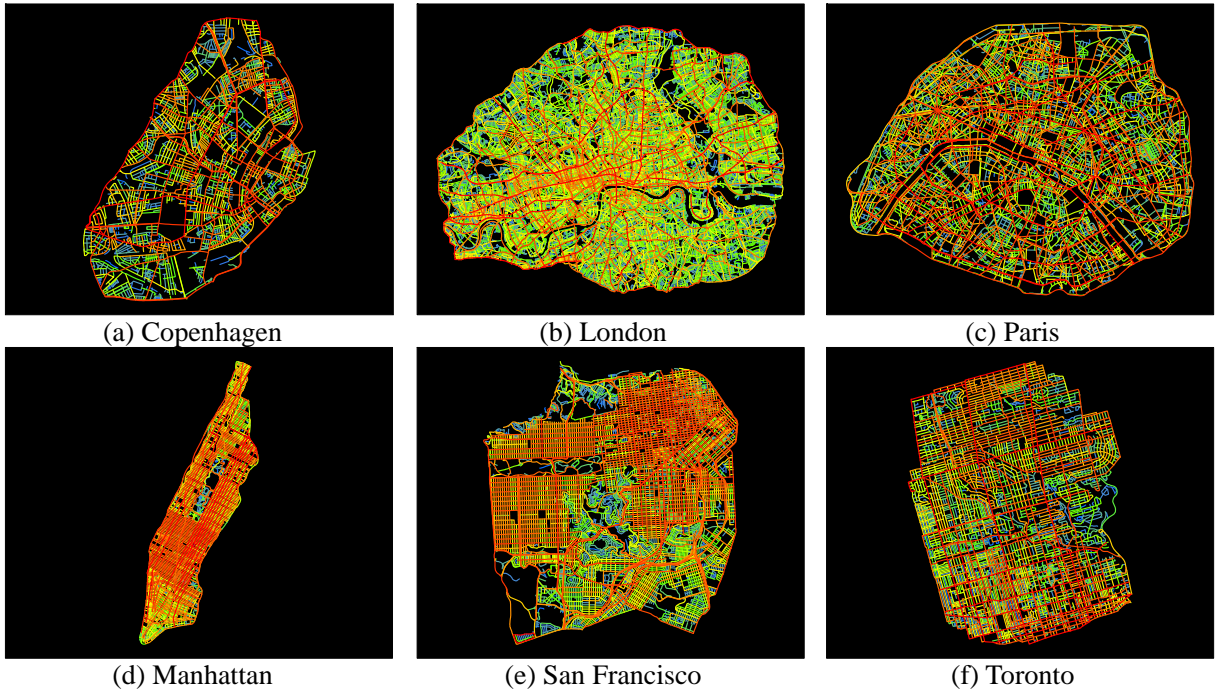

(a) Copenhagen (b) London (c) Paris

(d) Manhattan (e) San Francisco (f) Toronto

Figure 3: (Color online) Visualization of natural streets according to their local integration
(Note: A spectral color legend with smooth transition from blue to red is used for visualization; red lines indicate the highest local integration, and blue lines show the lowest local integration. In addition, the axial lines are drawn using the same transition order from blue to red, i.e., blue lines are first drawn, while red lines are drawn last)

Before generating the axial lines, we need to verify whether or not the two parameters, the base line length d and ratio x/d, follow a heavy tailed distribution including power law, exponential, lognormal, stretched exponential, and power law with a cutoff (Clauset et al. 2009). It is found that the two parameters exhibit a striking lognormal distribution as shown in Figure 4. This sets a prerequisite for using the head/tail division rule (Jiang and Liu 2010c) for the chopping process.

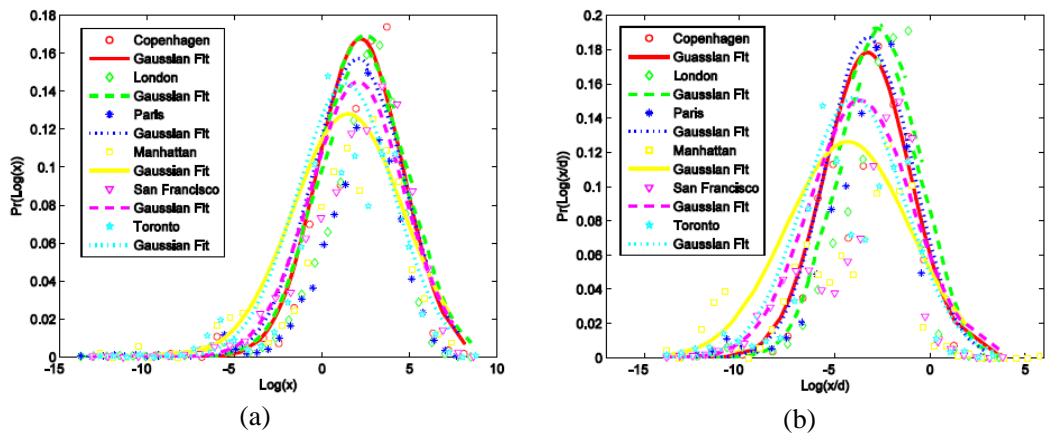

(a)  (b)

Figure 4: (Color online) Lognormal distribution of distance x and ratio x/d (c.f., Figure 1 for an illustration of the two parameters)

Next, based on the natural streets we automatically generate the axial lines according to the parameter settings provided in Table 1. The axial lines are visualized according to local integration using the same color legend and classification as for natural streets in Figure 3. It should be noted that the process of computing axial lines are pretty fast thanks to some efficient data structure. It cost only 25 seconds to generate 40 thousand lines for London, and a few seconds for other axial maps (Table 1).

Table 1: Number of axial lines generated with time cost and parameter settings

|                        | Copenhagen | London | Paris | Manhattan | San Francisco | Toronto |
|------------------------|-----------:|-------:|------:|----------:|--------------:|--------:|
| Number of axial lines  | 2382       | 42587  | 6846  | 1295      | 5067          | 3861    |
| Time cost (seconds)    | 3          | 25     | 5     | 2         | 5             | 3       |
| Mean(x) (meters)       | 37.2       | 28.5   | 31.6  | 46.7      | 55.8          | 35.6    |
| Mean(x/d) (%)          | 15         | 15     | 11    | 42        | 14            | 14      |

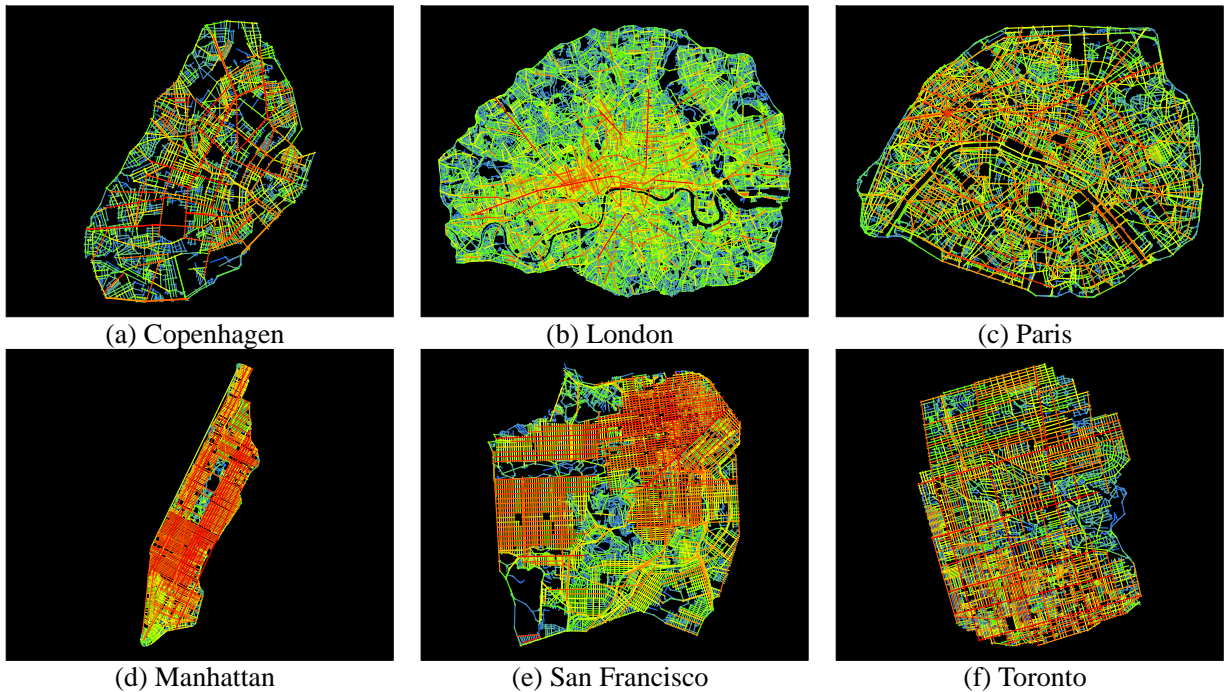

(a) Copenhagen  (b) London  (c) Paris
(d) Manhattan  (e) San Francisco  (f) Toronto

Figure 5: (Color online) Visualization of axial lines according to their local integration
(Note: A spectral color legend with smooth transition from blue to red is used for visualization; red lines indicate the highest local integration, while blue lines show the lowest local integration. In addition, the axial lines are drawn using the same transition order from blue to red, i.e., blue lines are drawn first and red lines are drawn last)

## 4. Results and discussion

Based on the above experiments, we find that the connectivity of both axial lines and natural streets follow a lognormal distribution, which is one of the heavy tailed distributions (Clauset, Shalizi and Newman 2009). This is in line with an earlier study (Jiang 2009), which claimed a power-law-like distribution for the London axial lines defined in the conventional way. The heavy tailed distribution implies that there are far more shorter (or less connected) streets than longer (or well connected) ones. It appears that the longest natural streets (in red) are much more common (Figure 3) than the longest axial lines (in red) (Figure 5). This is understandable since natural streets tend to aggregate more street segments than axial lines. Considering this, it might come natural to consider the axial lines as a better representation for capturing the underlying urban morphologies. This is because the fewer longest axial lines, the more memorable they are to city residents (Tomko, Winter and Claramunt 2008, Miller 1956). This point is applicable to the comparison between European cities and North American cities, i.e., there are fewer longest streets in European cities than in North American cities. We believe that the fewer longest streets in European cities tend to shape our mental maps better than the more longest streets in North American cities, simply because (1) there are many the longest streets in North American cities, and (2) the longest streets are somewhat with a similar length.

Let us take a more detailed look at the London axial map. There is already one manually drawn axial map based on the conventional definition of axial maps available at (http://eprints.ucl.ac.uk/1398/). Visual inspection of the two London axial maps, although the areas covered are very different (Figure 6), indicates that both maps capture the urban morphology well. For example, both the axial lines conventionally and newly defined exhibit the same lognormal distribution (Figure 7). More specifically, Oxford Street and those intersected with it constitute the core of the pattern, so they are shown in red in the axial maps. In the following, we will further make a comparison between the conventionally and newly defined axial lines from the point of view of capturing traffic flow.

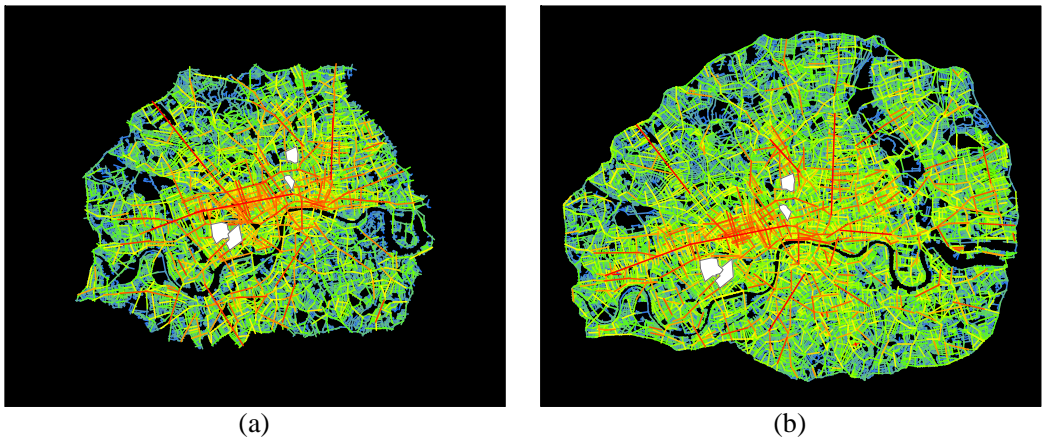

(a)                                                         (b)

Figure 6: (Color online) Visual comparison of London axial maps consisting of old axial lines (a) and new axial lines (b) (Note: the four patches indicate the four sites where observed pedestrian flow data are available for our comparison study)

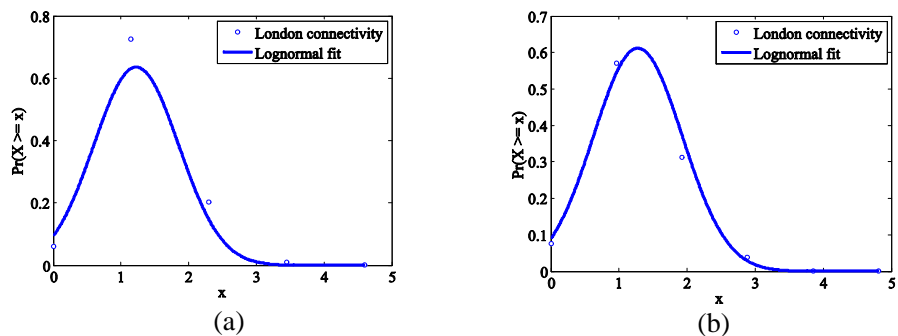

(a)                                                         (b)

Figure 7: Right-skewed lognormal distribution for both (a) old axial lines and (b) new axial lines of London

We adopt the observed pedestrian flow data available at (http://eprints.ucl.ac.uk/1398/), which have been used previously in a study (Hillier et al. 1993) to verify whether or not the new axial lines' space syntax metrics correlate with the traffic flow. There are a range of space syntax metrics among which local integration is supposed to be one of the best indicators of traffic flow. It should be noted that the observed data are of pedestrian flows captured in individual gates allocated in some street segments. This is not a perfect dataset, since not every street segment along a street has a gate. For example, some very long axial lines (or natural streets) covering or consisting of many street segments have only a couple of gates. However, it has been used as a benchmark dataset in the literature. Given the imperfectness of the data, we could not expect a very good correlation between location integration and traffic flow.

We manually pinpoint the individual gates in the new axial lines and the natural streets. It is a very tedious process. Eventually, we get the R square values between local integration and pedestrian flow for the three representations: new axial lines, old axial lines, and natural streets as shown in Table 2. In addition, we run a t-test indicating the correlation is statistically significant. From Table 2 and the t-test, we can conclude that the new axial lines capture well the pedestrian flow, at least as good as the old axial lines or natural streets. However, given that fact that generation of the new axial lines can be done automatically, the newly defined axial lines are a better representation than the old or conventional ones.

Table 2: Correlation coefficient (R square) between local integration and pedestrian flow

|  | Barnsbury | Clerkenwell | S. Kensington | Knightsbridge |
|---|---|---|---|---|
| New axial lines | 0.58 | 0.67 | 0.61 | 0.41 |
| Old axial lines* | 0.71 | 0.57 | 0.51 | 0.47 |
| Natural streets | 0.55 | 0.59 | 0.46 | 0.53 |

* from Hillier and Iida (2005)

We have illustrated that new axial lines and old axial lines are very similar at a global scale in terms of capturing the underlying structural patterns and traffic flow. However, we note that some new axial lines are better justified than the corresponding old axial lines; refer to Figure 8 for some highlighted axial lines in black. Obviously, the number of the new axial lines is (Figure 8b) fewer than that of the old axial lines (Figure 8a). From the criteria of the least number of lines, the new axial lines are better than the old axial lines. This can explain the fact shown in Table 2 that some correlation coefficients for the new axial lines are higher than that for the old axial lines and natural streets.

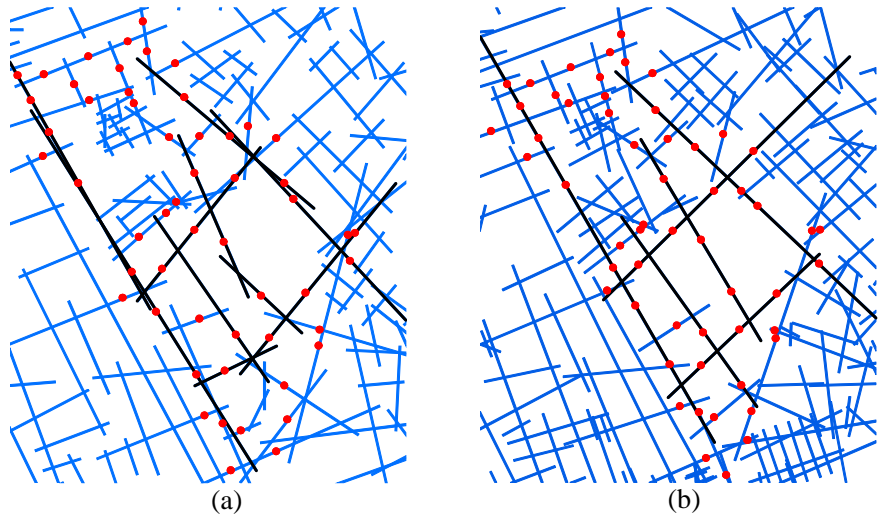

(a)  (b)

Figure 8: (Color online) Local view of the new axial lines (a) and the old axial lines (b) of the Clerkenwell site with pinpointed gates (red spots)

To this point, we have provided evidence that the newly defined axial lines can be an alternative representation to the conventional axial lines and to the natural streets. We have seen that the new axial lines could be a better alternative representation in capturing the underlying urban morphologies and consequently the traffic flow from the point of view of their conceptual justification and auto-generation.

## 5. Conclusion

This paper re-defined axial lines from the perspective of the walkability or drivability of streets or street networks rather than being based on the visibility between buildings or any spatial obstacles in cities. We have illustrated that this perspective makes better sense when generating axial lines at the city level. We define axial lines as the least number of relatively straight lines that are mutually intersected along individual natural streets. The new definition is less ambiguous compared to the conventional one. Based on this new perspective and definition, we develop an automatic solution to generating the axial lines for large cities. To our surprise, some thresholds related to curviness of natural streets can be determined by the head/tail division rule (Jiang and Liu 2010c), since they exhibit a heavy tailed distribution. We conduct a comparison study between the new axial lines and the old axial lines and between the new axial lines and natural streets. We have proved that the axial line representations are a powerful tool for urban studies just as Wagner (2007) indicated in his study on conventional axial lines.

We have illustrated through experiments and a comparison study that the new axial lines can capture the underlying urban morphologies just as good as the conventional axial lines and natural streets. Unlike the old axial lines, the new axial lines are defined without ambiguity. Both new and old axial lines seem based on the same principle of spatial perception, either in terms of visibility or drivability. On the other hand, the natural streets, like named streets identified by unique names, seem based on the spatial cognition for their definition or generation. We could remark that both axial lines (new or old) and streets (natural or named) are modeled from different perspectives, but they can all essentially capture the underlying urban morphology if they are correctly derived. In this respect, a ring road is represented as one street, but it is chopped into many axial lines. However, from the point of view of auto-generation, both the new axial lines and natural streets show a striking advantage to the old axial lines.

**Acknowledgement**
We thank the OSM community for voluntarily providing the data of the street networks. The solution developed in this paper has been implemented in Axwoman 5.0: http://fromto.hig.se/~bjg/Axwoman.htm.

**References:**
Clauset A., Shalizi C. R., and Newman M. E. J. (2009), Power-law distributions in empirical data, *SIAM Review*, 51, 661-703.
Douglas D. and Peucker T (1973), Algorithms for the reduction of the number of points required to represent a digitized line or its caricature, *The Canadian Cartographer*, 10(2), 112–122.
Figueiredo F., and Amorim L. (2005), Continuity lines in the axial system, in: *Proceedings of the Fifth Space Syntax International Symposium*, Delft University of Technology, 13 – 17 June, 2005, Delft.
Freeman L. C. (1979), Centrality in social networks: conceptual clarification, *Social Networks*, 1, 215 - 239.
Goodchild M. (2007), Citizens as sensors: The world of volunteered geography, *GeoJournal*, 69(4), 211-221.
Haklay M. and Weber P. (2008), OpenStreetMap: user-generated street maps, *IEEE Pervasive Computing*, 7(4), 12-18.
Hillier B. and Hanson J. (1984), *The Social Logic of Space*, Cambridge University Press: Cambridge.
Hillier B., Penn A., Hanson J., Grajewski T. and Xu J. (1993), Natural movement: configuration and attraction in urban pedestrian movement, *Environment and Planning B: Planning and Design*, 20, 29-66.
Hillier, B. (1996), *Space Is the Machine: a configurational theory of architecture*, Cambridge University Press: Cambridge.
Hillier, B. and Iida, S. (2005), Network and psychological effects in urban movement, In: Cohn, A.G. and Mark, D.M. (eds.), *Proceedings of the International Conference on Spatial Information Theory, COSIT 2005*, Ellicottsville, N.Y., U.S.A., September 14-18, 2005, Springer-Verlag: Berlin, 475-490.
Jacobs A. B. (1995), *Great Streets*, MIT Press: Cambridge, MA.
Jenks G. F. (1967), The data model concept in statistical mapping, *International Yearbook of Cartography*, 7, 186-190.
Jiang B. and Liu X. (2010c), Scaling of geographic space from the perspective of city and field blocks and using volunteered geographic information, http://arxiv.org/abs/1009.3635.
Jiang B. (2009), Ranking spaces for predicting human movement in an urban environment, *International Journal of Geographical Information Science*, 23.7, 823–837.
Jiang B. and Claramunt C. (2002), Integration of space syntax into GIS: new perspectives for urban morphology, *Transactions in GIS*, 6(3), 295-309.
Jiang B. and Claramunt C. (2004), Topological analysis of urban street networks, *Environment and Planning B: Planning and Design*, 31, 151- 162.

Jiang B. and Liu X. (2010a), Automatic generation of the axial lines of urban environments to capture what we perceive, *International Journal of Geographical Information Science*, 24.4, 545–558.

Jiang B. and Liu X. (2010b), Computing the fewest-turn map directions based on the connectivity of natural roads, *International Journal of Geographical Information Science*, x, xx-xx, Preprint: http://arxiv.org/abs/1003.3536

Jiang B., Zhao S., and Yin J. (2008), Self-organized natural roads for predicting traffic flow: a sensitivity study, *Journal of Statistical Mechanics: Theory and Experiment*, July, P07008.

Miller G. A. (1956), The magic number seven, plus or minus two: Some limits on our capacity for processing information, *The Psychological Review*, 63(2), 81–97.

Peponis J., Wineman J., Bafna S., Rashid M., and Kim S. H. (1998), On the generation of linear representations of spatial configuration, *Environment and Planning B: Planning and Design*, 25, 559 – 576.

Ramer U. (1972), An iterative procedure for the polygonal approximation of plane curves, *Computer Graphics and Image Processing*, 1(3), 244–256.

Ratti C. (2004), Space syntax: some inconsistencies, *Environment and Planning B: Planning and Design*, 31, 487 – 499.

Thomson R. C. (2003), Bending the axial line: smoothly continuous road centre-line segments as a basis for road network analysis, *Proceedings of the 4th International space syntax symposium*, London 17-19 June 2003.

Tomko M., Winter S. and Claramunt C. (2008), Experiential hierarchies of streets, *Computers, Environment and Urban Systems*, 32(1), 41-52.

Turner A., Penn A., and Hillier B. (2005), An algorithmic definition of the axial map, *Environment and Planning B: Planning and Design*, 32, 425 – 444.

Wagner R. (2007), On the metric, topological and functional structures of urban networks, *Physica A*, 387(8-9), 2120 - 2132.